# TD3 Based Collision Free Motion Planning for Robot Navigation

Hao Liu [1],[*]
Northeastern University
Shenyang, China
liuhao@stumail.neu.edu.cn

Yi Shen[2]
University of Michigan, Ann Arbor
USA
shenrsc@umich.edu

Chang Zhou[3]
Columbia University
USA
mmchang042929@gmail.com

Yuelin Zou[4]
Columbia University
USA
yz4198@columbia.edu

Zijun Gao[5]
Northeastern University
USA
gao.zij@northeastern.edu

Qi Wang[6]
Northeastern University
USA
bjwq2019@gmail.com

*Abstract*—This paper addresses the challenge of collision-free motion planning in automated navigation within complex environments. Utilizing advancements in Deep Reinforcement Learning (DRL) and sensor technologies like LiDAR, we propose the TD3-DWA algorithm, an innovative fusion of the traditional Dynamic Window Approach (DWA) with the Twin Delayed Deep Deterministic Policy Gradient (TD3). This hybrid algorithm enhances the efficiency of robotic path planning by optimizing the sampling interval parameters of DWA to effectively navigate around both static and dynamic obstacles. The performance of the TD3-DWA algorithm is validated through various simulation experiments, demonstrating its potential to significantly improve the reliability and safety of autonomous navigation systems.

## I. INTRODUCTION

In the context of rapid development of modern technology, the role of collision-free motion planning for robots is increasingly critical in the realm of automated navigation, particularly within the complex urban traffic systems [1]. The introduction of Deep Reinforcement Learning (DRL) has been extensively utilized to tackle these challenging dynamic scenarios [2][3]. Research by reference [4] has enhanced the positioning accuracy of vehicles in complex environments through particle filter SLAM technology, providing a technical foundation for the practical application of automated navigation systems.

Reference [5] highlighted the evolution from simple assistive devices to advanced manufacturing collaborative robot groups, underscoring the urgent need to develop robust adaptive learning systems in robotics. This progress emphasizes the importance of developing sophisticated motion planning algorithms that can meet the demands of robotics [6].

Reference [7] explored the use of Bidirectional Gated Recurrent Units for sentiment analysis, revealing the potential of deep learning models to handle complex datasets, a fundamental capability for robots operating in unpredictable environments. This ability to interpret and respond to environmental cues forms the basis for developing autonomous systems that can navigate without preset instructions [8].

Effectively handling the uncertainties and dynamics of real-world scenarios is a crucial issue [9]. Reference [10] provides an in-depth study of the Monte Carlo tree search algorithm and its performance in games, providing methodological insights for stochastic planning techniques in robotic.

Safety considerations are also crucial in the deployment of autonomous systems. Reference [11] explored hardware-based Spectre attack detection techniques, emphasizing the importance of integrating robust security measures into systems performing complex algorithms to ensure the integrity and reliability of autonomous navigation systems.

Reference [12] enhanced fault detection in time-series traffic sensor data through symmetric contrastive learning. Their approach has improved the accuracy of sensor data analysis, which is vital for the real-time processing capabilities required for autonomous navigation[13]. Reference [14] developed a learning technique for 3D point cloud data, enhancing robots' spatial understanding and environmental interaction. This has bolstered their ability to navigate complex terrains by more effectively identifying and avoiding obstacles [15].

As robotics technologies continue to evolve, robots are being equipped with increasingly sophisticated perceptual capabilities. Today, a variety of sensors, such as LiDAR, are widely used in robotics[16], and the recent advancements in reinforcement learning have enabled robots to move beyond traditional methodologies[17]. The real world is replete with a variety of unknown obstacles, both dynamic and static[18][19]. While static obstacle avoidance has been largely addressed, navigating dynamic environments remains an unresolved challenge[20]. Furthermore, robot trajectory planning involves multiple evaluation factors, and generating practical trajectories is a critical issue that needs addressing. Reference [21] introduced an adaptive dynamic window approach alongside a novel deep convolutional neural network, which effectively addresses the issue of parameter selection in the DWA. However, it does not resolve the inherent issues associated with the DWA algorithm.

This paper proposes the TD3-DWA algorithm, which solves the problem of optimizing the sampling interval parameters of the traditional DWA algorithm. This algorithm generates directly usable local trajectories, and combines the DWA algorithm's own planning primitives and cost functions to generate trajectories that can avoid dynamic obstacles.

## II. METHODOLOGY

### A. Dynamic Window Approach (DWA)

The DWA algorithm is a relatively conventional local path planning algorithm. Its main steps include: first sampling the speed of the smart car according to the characteristics of the smart car and the environment; then predicting the trajectory based on the sampled speed; finally using the evaluation function to select the optimal trajectories for path planning. The kinematics model of the smart car is (1):

$$\begin{cases} \dot{x}_i = v_i \cos\theta_i \\ \dot{y}_i = v_i \sin\theta_i \\ \dot{\theta}_i = \omega_i \end{cases} \quad (1)$$

The pose of the robot is determined by the location and movement direction of the robot in space. During the robot speed sampling process, assuming that its maximum linear acceleration is $a$ and the robot speed in the current state is $v_0$, the linear speed change within unit time $dt$ is $v = v_0 + a*dt$. In actual operation, the robot will not have a negative linear speed, and it is impossible for the linear speed to increase infinitely. According to the actual driving capability of the robot, the minimum linear speed is 0 and the maximum linear speed is $v_{\max}$. Since the robot does not always maintain a linear acceleration of a during $dt$ time, but floats between 0 and $a$, the linear velocity range of the robot within $dt$ time is :

$$v \in \{v_0 - a*dt, v_0, v_0 + a*dt\} \quad (2)$$

$$\omega \in \{\omega_0 - a*dt, \omega_0, \omega_0 + a*dt\} \quad (3)$$

Once the velocity space has been sampled during prediction time, a cost function can be used to evaluate the goodness of the predicted trajectory. The evaluation function of the DWA algorithm usually needs to be set according to the actual situation, mainly including the cost of target distance, speed, obstacles and azimuth angle. In this article, the evaluation function of the DWA algorithm is set according to (4), that is, the smaller the total cost, the better the path planning process.

$$G(v,w) = \sigma(\alpha * dist(v,w) + \beta * heading(v,w) + \gamma * vel(v,w)) \quad (4)$$

$dist(v,w)$ is the distance evaluation function between the predicted path and each obstacle, $heading(v,w)$ is the direction evaluation function between the robot and the target point, $vel(v,w)$ is the speed evaluation function of the robot. $\sigma$ is the normalization function. $\alpha$, $\beta$, $\gamma$ are the weights of three evaluation function factors respectively.

### B. TD3-DWA algorithm

In the traditional DWA planning method, the sampled values of linear and angular velocities at each time step within the same trajectory are treated as constants. This limitation restricts the potential reachability of the vehicle within the prediction domain, posing challenges in finding reasonable solutions for robot trajectories in potentially feasible scenarios. While this constraint may not significantly impact planning in structured environments, it becomes more pronounced when scaling up to scenarios involving higher speeds and longer forecast periods.

Considering the computational costs associated with precise calculations, the sampling interval is inevitably constrained. Consequently, this broad sampling range further escalates the time complexity of robot planning, diminishing planning efficiency and increasing the risk of missing feasible solutions.

The fixed sampling values within the same trajectory and constant sampling intervals across different trajectories serve as both the foundation for ensuring real-time performance and a limitation on the effectiveness of DWA planning. Introducing dynamic sampling values or intervals significantly increases computational costs as the prediction horizon extends. Thus, to strike a balance between real-time performance and planning efficiency, an autonomous planning method based on Reinforcement Learning is proposed to enhance exploration within a dynamic sampling framework.

By incorporating a reinforcement learning module, the proposed method utilizes the full range of sampling possibilities within the dynamic window to continuously generate competitive planning trajectories with independent online planning capabilities. This comprehensive planning approach across sample points along the prediction domain yields a versatile and feasible trajectory sequence online, serving as an auxiliary competitive trajectory.

As previously mentioned, we approach the control of linear velocity and angular velocity within the framework of reinforcement learning, framing the problem as a Markov decision process (MDP) characterized by tuples $(S, A, P, R, \gamma)$. In this setup, an agent interacts with the environment by starting from an initial state $s_0$ drawn from a predetermined distribution $p(s_0)$. Subsequently, it cyclically selects actions $a_t \in A$ from a general state $s_t \in S$, transitioning to a new state $s_{t+1}$ with a transition probability $P(s_{t+1} | s_t, a_t)$, and receiving the corresponding reward $rt = R(s_t, a_t)$.

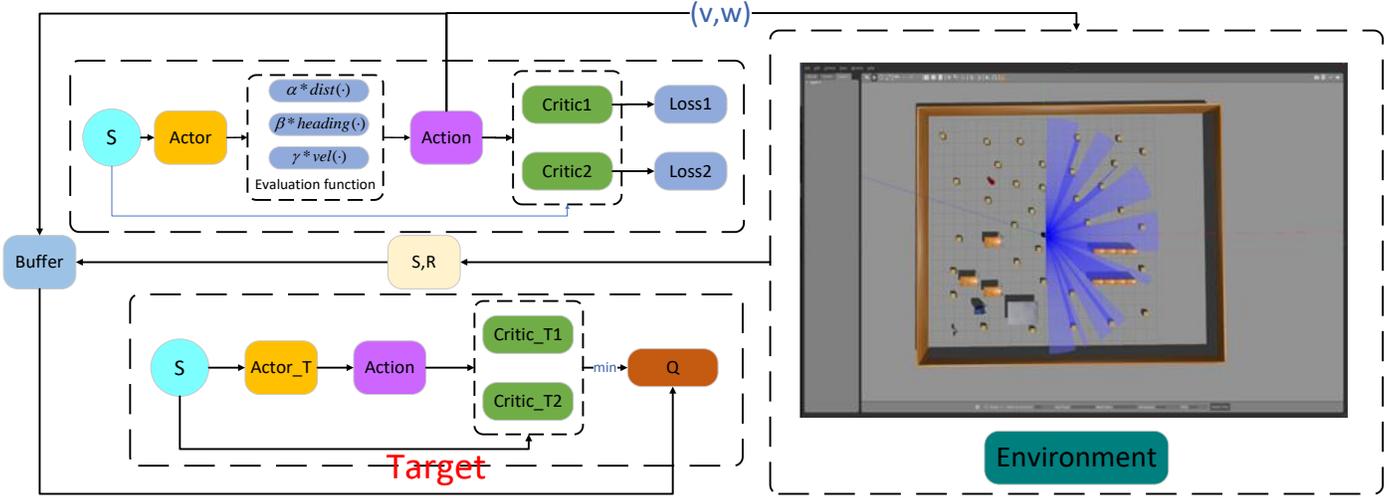

Fig 1. Structure of TD3-DWA algorithm

The TD3 (Twin Delayed Deep Deterministic Policy Gradient) algorithm is a reinforcement learning algorithm used to solve continuous action space problems, proposed by Fujimoto et al. in 2018. It is based on the DDPG (Deep Deterministic Policy Gradient) algorithm and introduces three major improvements to solve some problems that DDPG may encounter during the training process, especially overestimation when estimating the Q value.

The whole process is shown in the Fig. 1. TD3 uses two independent Q-networks (critic) to reduce deviation in value function estimates. During training, TD3 selects smaller Q values for gradient updates in both networks, thereby reducing the impact of overestimation. In TD3, the policy network (actor) is updated less frequently than the Q-network. This means that the policy network is only updated once every few times the Q-network is updated. This delayed update strategy can ensure that the policy network is updated based on a relatively stable value function, thereby improving the stability of learning.

By integrating the TD3 network with the DWA algorithm, the trajectory generation process can be enhanced. Instead of generating multiple reference trajectories through sampling as in the traditional DWA algorithm, the $(v, w)$ velocities output by the TD3 network can be used directly. Traditionally, DWA outputs an instantaneous evaluation based on the assumption that motion remains constant and predictable over several forward time steps. However, in reinforcement learning, cumulative rewards are derived from continuous interaction with the simulated environment throughout an episode, which more accurately mirrors the dynamics in actual physical models.

We equip each differential robot with lidar to sense the local environment. For each moment, a position of the obstacle relative to the odometry coordinate system of the differential robot can be obtained. By dividing the lidar into laser beams at fixed intervals, the number of obstacles perceived by the robot at each moment can be fixed, thereby combining dimensionality reduction with the obstacles.

## III. EXPERIMENTS

All experiments were performed on a computer which is equipped with an Intel(R) Core(TM) i7-7700HQ CPU @ 2.80GHz and an NVIDIA GPU. GeForceGTX1080GPU. All experiments are based on the ubuntu20.04 operating system and ROS, and the physical simulation is performed in Gazebo.

As is shown in Fig.2, we verified the effectiveness of the TD3-DWA algorithm in two simulated scenarios, one based on an environment with only static obstacles and the other with dynamic pedestrians. The simulation environment consists of a closed space, approximately 10*15 meters in size, with a large number of static obstacles and dynamic obstacles. The robot uses a differential drive mechanism, and the maximum linear speed and angular speed of the robot are set to 1.2 m/s and 1.57 rad/s respectively.

The $\varepsilon$-greedy exploration strategy is defined by the initial value $\varepsilon_0 = 1.0$, the decay $\gamma_\varepsilon = 0.992$, and the minimum value $\varepsilon_{min} = 0.05$ for random action sampling. The agent is trained for 3300 epochs, including $T = 300$ max steps. The starting pose of the robot changes every 20 episodes to ensure a good level of exploration and eventual promotion.

In the TABLE Ⅰ, we compare the effects of four algorithms, namely TD3-DWA algorithm, DQN-DWA algorithm, DWA algorithm and TEB algorithm. In the TABLE Ⅰ, in terms of collisions, the TD3-DWA algorithm only experienced two collisions in 100 average tests, which is significantly better than other algorithms. In terms of average path length, the

DQN-DWA algorithm performs best, followed by the TD3-DWA algorithm. In terms of path time consumption, the TD3-DWA algorithm still performs best. This is because the TD3-DWA algorithm performs best when selecting actions. The selected actions can better maintain the smoothness of the speed, and the entire path can be traveled smoothly during the entire planning process, while the DQN-DWA algorithm encountered multiple stalls. The performance of the traditional DWA algorithm and TEB algorithm is significantly weaker than the above two methods. Overall, the TD3-DWA algorithm has the best overall performance. In Table II, the test environment is 25*25 meters and the number of obstacles remains unchanged. The TD3-DWA algorithm outperforms other algorithms in all indicators and is completely collision-free.

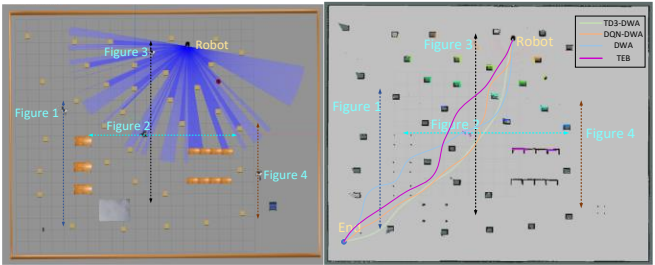

Fig 2. Dynamic environment planning

TABLE I.  PERFORMANCE IN 10*15 METERS

| Method | Collision | Avg. Time (s) | Avg. Path Length (m) |
|---|---|---|---|
| TD3-DWA | **2/100** | **11.6** | 11.6 |
| DQN-DWA | 3/100 | 14.5 | **10.3** |
| DWA | 14/100 | 18.4 | 13.7 |
| TEB | 11/100 | 21.9 | 15.8 |

TABLE II.  PERFORMANCE IN 25*25 METERS

| Method | Collision | Avg. Time (s) | Avg. Path Length (m) |
|---|---|---|---|
| TD3-DWA | **0/100** | **22.3** | **20.4** |
| DQN-DWA | 1/100 | 29.5 | 23.8 |
| DWA | 8/100 | 26.1 | 27.4 |
| TEB | 8/100 | 30.4 | 25.2 |

## IV. CONCLUSIONS

To address the challenges of path planning and collision avoidance in unmanned vehicles, traditional DWA often falls short, exhibiting low efficiency in path planning, suboptimal path outcomes, susceptibility to local optima, and challenges in selecting weight coefficients for objective functions. This paper proposes an improved solution by integrating DWA with TD3 algorithm, aiming to enhance real-time collision avoidance in dynamic settings. The effectiveness of this hybrid algorithm is demonstrated through various comparative experiments in a simulation environment.